# From operculum and body tail movements to different coupling of physical activity and respiratory frequency in farmed gilthead sea bream and European sea bass. Insights on aquaculture biosensing

Miguel A. Ferrer , Josep A. Calduch-Giner , Moises Díaz, , Javier Sosa , Enrique Rosell-Moll , Judith Santana Abril , Graciela Santana Sosa , Tomás Bautista Delgado , Cristina Carmona , Juan Antonio Martos-Sitcha , Enric Cabruja , Juan Manuel Afonso , Aurelio Vega , Manuel Lozano , Juan Antonio Montiel-Nelson , Jaume Pérez-Sánchez.

## Abstract

The AEFishBIT tri-axial accelerometer was externally attached to the operculum to assess the divergent activity and respiratory patterns of two marine farmed fish, the gilthead sea bream (Sparus aurata) and European sea bass (Dicentrarchus labrax). Analysis of raw data from exercised fish highlighted the large amplitude of operculum aperture and body tail movements in European sea bass, which were overall more stable at low-medium exercise intensity levels. Cosinor analysis in free-swimming fish (on-board data processing) highlighted a pronounced daily rhythmicity of locomotor activity and respiratory frequency in both gilthead sea bream and European sea bass. Acrophases of activity and respiration were coupled in gilthead sea bream, acting feeding time (once daily at 11:00 h) as a main synchronizing factor. By contrast, locomotor activity and respiratory frequency were out of phase in European sea bass with activity acrophase on early morning and respiration acrophase on the afternoon. The daily range of activity and respiration variation was also higher in European sea bass, probably as part of the adaptation of this fish species to act as a fast swimming predator. In any case, lower locomotor activity and enhanced respiration were associated with larger body weight in both fish species. This agrees with the notion that selection for fast growth in farming conditions is accompanied by a lower activity profile, which may favor an efficient feed conversion for growth purposes. Therefore, the use of behavioral monitoring is becoming a reliable and large-scale promising tool for selecting more efficient farmed fish, allowing researchers and farmers to establish stricter criteria of welfare for more sustainable and ethical fish production.

## 1. Introduction

Accelerometers are widely used to assess physical activity in public health (Vale et al., 2015) as they provide reliable measurements of energy expenditure and time spent in different activity conditions (Crouter et al., 2018). Certainly, activity recognition by means of specific algorithms allows the risk assessment of sedentary lifestyle and overweight in aged people (Taylor et al., 2014) and children (Duncan et al., 2016; Roscoe et al., 2019), which enables the use of accelerometer records for extracting quantitative measures of biological age (Pyrkov et al., 2018). Since the late 1990s, researchers have also employed portable accelerometers to investigate energy expenditure, activity patterns and the postural behavior of livestock, companion animals, free-ranging species, laboratory animals and zoo-housed species (Brown et al., 2013; Whitham and Miller, 2016). However, it is important to certify that these devices do not negatively impact the animals and, hence, skew the data. Thus, important research efforts are focused on how and where the device is attached. Common attachment methods include collars, anklets, harnesses and clamps, and the placement of the device determines the type of behavior that can be monitored (Brown et al., 2013). Furthermore, to consider whether the subject or conspecifics can remove the device is a key factor (Rothwell et al., 2011), and whether color, mass or shape affect the animal behavior, that limits the functional value of the registered data, is also of importance (Wilson et al., 2008). Ruminants are, however, a case of high success and a number of studies clearly indicate that feeding behavior (Alvarenga et al., 2016; Rayas-Amor et al., 2017), rumen mobility (Michie et al., 2017; Hamilton et al., 2019) or positive affective states affecting diseases and other welfare concerns (de Oliveira and Keeling, 2018) are measurable by accelerometers, contributing to improve animal welfare and productivity.

Like in terrestrial livestock, the biosensor technology has the potential to revolutionize the aquaculture industry (Andrewartha et al., 2016; Endo and Wu, 2019; Rajee and Alicia, 2019), but the state-of-theart of micro-systems provide limited real-time access to telemetry data (Føre et al., 2018; Hassan et al., 2019). Size and energy autonomy are also obvious limitations, and the choice of tagging method (external, attachment, surgical implantation), operational mode (stand-alone vs. wireless systems) and telemetry technology (e.g. radio-transmitters, acoustic transmitters, pop-up satellite archival tags) ultimately depends on life species, life stage and research question (Thorstad et al., 2013; Jepsen et al., 2015). Thus, small and light devices working in standalone mode appear especially suitable for quickly tracking challenged fish at specific time windows, allowing farmers and breeders to orientate selective breeding towards more robust and efficient fish or improve culture conditions for a more sustainable and ethical production. These are the criteria used within the AQUAEXCEL$^{2020}$ EU project for the design of AEFishBIT, a patented (P201830305), stand-alone, small and light (1 g) motion embedded-microsystem based in a tri-axial accelerometer that is externally attached to the operculum to monitor physical activity by mapping accelerations in x- and y-axes, while operculum beats (z-axis) serve as a measurement of respiratory frequency (Martos-Sitcha et al., 2019). The accuracy of on-board algorithms was calibrated in swim metabolic chambers, and accelerometer outputs of exercised gilthead sea bream (Sparus aurata) and European sea bass (Dicentrarchus labrax) juveniles correlated with data on swimming speed and oxygen consumption. However, these two economically important marine farmed fish exhibit different locomotor capabilities, and we aimed to provide new insights into their divergent patterns of activity and energy partitioning between growth and locomotor activities. To pursue this global aim, raw data from forced exercised fish in metabolic chambers (15 min) were retrieved to analyze the frequency and amplitude of operculum and body tail movements. Additionally, new post-processed data using on-board algorithms were obtained over extended recording periods (2 days) to assess the behavioral patterns of free-swimming fish in rearing tanks. Such approach reinforced the different adaptive strategies of gilthead sea bream and European sea bass that primarily arise from changes in body shape and specialized movements, but also from the different coupling on a daily basis of physical activity and respiratory frequency.

## 2. Materials and methods

### 2.1. Swim tunnel and raw data download

Data from exercised juveniles of gilthead sea bream (n = 18) and European sea bass (n = 15) in a swim tunnel respirometer (Loligo® Systems, Viborg, Denmark) were retrieved from Martos-Sitcha et al. (2019) for raw data analyses. Briefly, fish were exercised at three increasing speeds (1, 2 and 3 BL/s) lasting 5 min each consecutive period. Accelerometers were programmed for the acquisition of data sets for 2 min at each swimming speed at a sampling period of 100 Hz.

After testing, fish were removed from the tunnel and the device was pluggedout for data downloading and raw data off-line post-processing.

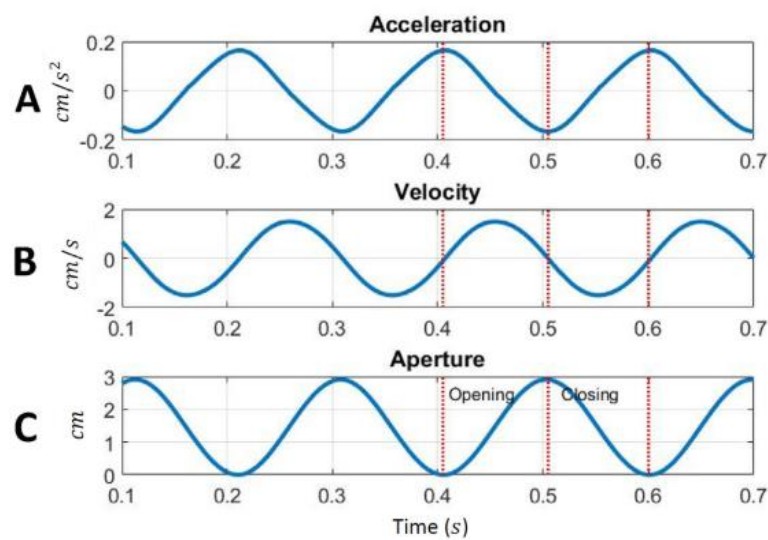

*Fig. 1. Synthetic example of aperture distance, velocity and acceleration during operculum movement. Acceleration is the derivate of the velocity and velocity the derivate of distance. A. Operculum closing (aperture = 0) and opening (maximum aperture) along time. B. Operculum velocity. C. Operculum acceleration. During operculum opening, velocity is positive and acceleration goes from positive to negative. During operculum closing closing, velocity is negative and acceleration goes from negative to positive.*

### 2.2. Raw data processing

The signal from the z-axis, that records the operculum opening and closing, was numerically integrated to assess the velocity of the operculum movement. This integration minimized the influence of other movements such as lateral body movements or angular velocity:

$$V_{zr}(t) = \int_0^t a_z(t)dt$$

To remove the trend of $V_{zr}(t)$, it was high pass filtered by a filter cutoff frequency of $f_c = 1$ Hz that allows breathing frequency pass through. The high pass filter was carried out in two steps: a) $V_{zr}(t)$ was low pass filtered by 1 Hz cut-off filter obtaining $V_{zlp}(t)$, and b) the detrended velocity of the operculum was estimated as $V_z(t) = V_{zr}(t) - V_{zlp}(t)$. The distance run by the operculum can be approached by integrating $V_z(t)$. Details and examples of processed signals and raw data are provided as Supplemental file S1.

Operculum opening can be defined as the distance increase from zero to maximum aperture, as illustrated in the synthetic example of Fig. 1A. The velocity was zero at the beginning and at the end of the opening, so the velocity increased and decreased in a bell-shaped way, as exemplified in Fig. 1B. In consequence, the acceleration (Fig. 1C) started high and positive during the first phase of the operculum opening and decreased to negative values, whereas the operculum started to stop at the second phase of the operculum opening.

Fig. 1 is consistent with the kinematic theory of the rapid movement that establishes a lognormality principle, which allows modelling the output velocity of a complex neuromuscular system through an overlapped sequence of lognormal functions. Based on the application of this principle, rapid human hand movements have been modeled (Plamondon et al., 2014; Diaz et al., 2015; Duval et al., 2015), and the same approach was used herein for modelling the operculum movement. After that, elemental opening and closing movements are represented as the following sequence of lognormal-shape velocity profiles (or lognormals):

$$v(t) = \sum_i v_i(t; t_{oi}; \mu_i; \sigma_i^2)$$

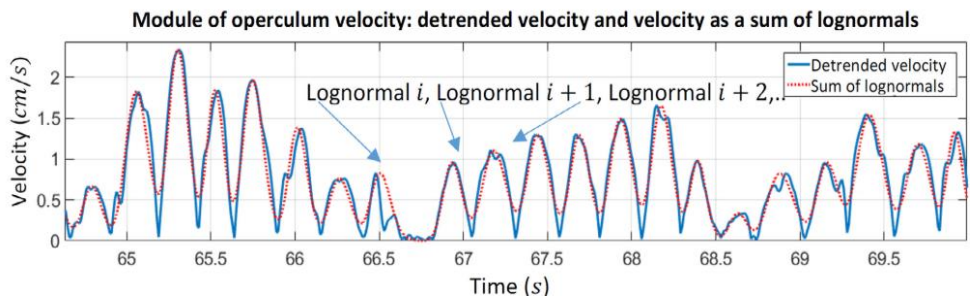

*Fig. 2. Module of the detrended velocity decomposed as a sum of lognormals.*

Being each velocity profile

$$v_i(t; t_{oi}; \mu_i; \sigma_i^2) = \frac{D_i}{\sigma_i\sqrt{2\pi}\,(t-t_{oi})} = \exp\left(-\frac{\ln(t-t_{oi})-\mu_i]^2}{2\sigma_i^2}\right)$$

where t is time; $t_{oi}$, time of movement occurrence; $D_i$, area of the velocity; $\mu_i$, time delay; $\sigma_i$, response time of each lognormal of the sequence. Both µi and σi are expressed on a logarithmic time scale. Thus, basic movements (opening or closing) of the operculum can be parameterized with $t_{oi}$, $D_i$, $\mu_i$ and $\sigma_i$ as exemplified in Fig. 2 meterized with $t_{oi}$, $D_i$, $\mu_i$ and $\sigma_i$ as exemplified in Fig. 2. Data measurements of operculum velocity yielded other parameters derived from the lognormal function:

1. D or area of the lognormal (distance run by the lognormal).

2. μ or lognormal time delay.

3. σ or lognormal response time.

Additional parameters were obtained from the lognormal shape (Fig. 3):

4. Temporal width (s): $t_3 - t_1$, time interval of a single movement.

5. Rise time (s): $t_2 - t_1$, time span of positive acceleration (velocity increment).

6. Drop time (s): $t_3 - t_2$, time span of negative acceleration (velocity decrement).

7. Lognormal skew, which estimates the velocity shape asymmetry. A skewness value between −0.5 and 0.5 means a fairly symmetrical movement. A skewness value > 0.5 indicates that the first phase of the movement is quicker than the second.

8. Kurtosis of the lognormal to estimate the "tailedness" of the velocity shape. A value of 3 means a Gaussian shape. Values larger than 3 indicate a leptokurtic movement, with extended tails and sharper and higher peaks.

The values of these eight parameters were obtained using iDeLog software, which includes recent improvements in the Sigma-Lognormal model (Ferrer et al., 2018).

A similar procedure was conducted to model the velocity of the x and y-axis, which defined the body tail movement. First, the accelerometer signals for both axes were numerically integrated:

$$V_{xr}(t) = \int_0^t a_x(t)dt$$

$$V_{yr}(t) = \int_0^t a_y(t)dt$$

Such velocities were detrended with a similar procedure than above obtaining $v_x(t)$ and $v_y(t)$. Then, the velocity of the body tail movement was estimated as $v_b = \sqrt{v_x^2(t) + v_y^2}$. This velocity was decomposed in a family of lognormals functions, which allows the extraction of the eight parameters mentioned above.

### 2.3. Free-swimming monitoring by means of on-board algorithms

To assess the free-swimming activity patterns of cultured fish, AEFishBIT measures were obtained from 3-year old gilthead sea bream (917.0 ± 37.2 g, n = 8) and European sea bass (645.1 ± 49.3 g body weight, n = 8) reared in 3,000 L tanks (8–10 kg/m3 ) at the indoor experimental facilities of Institute of Aquaculture Torre de la Sal (IATSCSIC) under natural photoperiod and temperature conditions (40° 5′N; 0° 10′E). Fish were fed once daily at 11:00 h, being overnight fasted the day of device tagging. The devices used for data recording were adequately packaged in silicone for water protection (14 × 7 × 7 mm), with a resulting weight in air of 1.1 g. The devices were externally attached to the operculum using metal Monel piercing fish tags (National Band & Tag Company, Newport, KY, United States) with a flexible heat shrink polyethylene tube (Eventronic, Shenzen, China) that is able to easily fit the device as shown at https://vimeo.com/ 325943543. In skilled hands, the entire application procedure took < 30 s per fish and no pathological signs of hemorrhage or tissue damage were found after 2–3 weeks of device tagging. Devices were programmed for on-board calculation of respiratory frequency and physical activity over 2 min time windows each 15 min along two consecutive days. Fish remained unfed over the recording time, and the devices were retrieved for downloading the processed data just after data recording was completed. For each device, clock time drift was previously estimated for post-processing synchronization. This time drift was established to be constant for a given device in a temperature range of 4–30 °C.

*2.4. Statistical análisis*

Interspecies comparisons for raw data results derived from operculum and body tail movements was conducted through the nonparametric Mann-Whithney U test, using the Matlab statistical toolbox. Analysis of on-board processed data of physical activity and respiratory frequency was assessed through Student's t-test and Pearson coefficients using the Sigmaplot suite (Systat Software Inc., Chicago USA). The daily rhythmicity of this time series analysis was further analyzed using a simple cosinor model (Refinetti et al., 2007).

*2.5. Ethics statement*

   No mortalities were observed during fish manipulation and experimental procedures. All procedures described herein were approved by the Ethics and Animal Welfare Committee of IATS-CSIC and carried out according to the National (Royal Decree RD53/2013) and the current EU legislation (2010/63/EU) on the handling of experimental fish.

**3. Results**

*3.1. Outlook of operculum movements*

   Lognormals derived from z-axis signal were aligned by fixing $t_0 = 0$. For each species and swim speed, the average shape of alternate velocity lognormals (i.e. shape comparison of operculum opening and closing movements) was the same (Supplemental Fig. S1), and both movements were considered as equivalent for calculations. Averaged lognormals are summarized in Table 1. For a given speed, the averaged values of temporal widths ($t_3 - t_1$, time for operculum opening or closing) were consistently lower in gilthead sea bream than European sea bass. In both species, rise time ($t_2 - t_1$, related with the agonist's muscle of the movement) was lower than drop time ($t_3 - t_2$, related with the antagonist's muscle) at any swim speed, indicating that the agonist muscle of the movement was always quicker than the antagonist one. Positive values of skewness confirmed this fact, as it indicated that the velocity profile was skewed towards left with a tail on the right side. Regarding the area of the lognormal (aperture of the operculum), D, it was always larger in European sea bass than in gilthead sea bream at any given velocity, indicating a greater operculum aperture in European sea bass. Overall, these parameters pointed out that this species breathes fewer times per second than gilthead sea bream, also showing a larger aperture of the operculum with a higher ability to keep a stable movement with changes in swim speed.

   With the increase of swimming speed, temporal width decreased in both species, with a concomitant decrease of rise and drop times, as well as σ and kurtosis. This increase of the respiratory frequency with increasing swimming speed was accompanied by an increase of D and a lowering of skewness, so the velocity profiles became more symmetric (Fig. 4). These findings were indicative of an increase of the respiratory frequency and a larger operculum opening with increasing swimming speed in both species.

*3.2. Outlook of body tail movements*

   Similarly to operculum movement analysis, all lognormals were aligned by fixing $t_0 = 0$. Averaged lognormals of gilthead sea bream and European sea bass body tail movements at different swimming speeds are summarized in Table 2. With the increase of water speed, temporal width decreased in both species, but it was sharper in $t_3 - t_2$ than in $t_2 - t_1$, which again suggests that the antagonist muscle become more and more active as swimming speed increases. It was also noticeable that D increased with swim speed in the case of gilthead sea bream, but it remained quite unaltered in European sea bass. This would imply that in order to compensate the increasing speed, European sea bass would increase the frequency of the movement, while gilthead sea bream would need to increase both frequency and amplitude of the body tail movement.

*3.3. Free-swimming temporal patterns of physical activity and respiration*

   Recorded data from incomplete light and dark phases, corresponding to the beginning and the end of the experimental period, were excluded to avoid any temporal bias. Thus, the analyzed period for all implanted individuals comprised two complete dark and one complete light phase. Mean values over time were extracted for preliminary analysis (Supplemental Fig. S2), and they were quite similar to the calculated mesor by means of cosinor analysis. For a given species, mesor values remained fairly constant among individuals, but pronounced differences were found between gilthead sea bream and European sea bass. Hence, the retrieved physical activity of gilthead sea bream was significantly higher than that of European sea bass ($0.080 \pm 0.006$ vs. $0.057 \pm 0.002$, P < 0.01). Conversely, respiratory frequency was significantly lower in European sea bass ($1.57 \pm 0.04$ vs. $1.73 \pm 0.04$, P < 0.05). In both fish species, correlation analysis of individual body weight with their own physical activity and respiratory frequency resulted in negative and positive correlations, respectively (Table 3). This stated that, for a given species, larger fish showed a lower physical activity and a higher respiratory frequency than their smaller congeners of the same age. Cosinor analysis of AEFishBIT recorded data showed clear daily rhythmic variations (Fig. 5).

For both fish species, the acrophase (time of peak value) of physical activity occurred few hours after the light onset, acting the pre-existing feeding time at 11:00 h as a main synchronizing factor (Fig. 5A, D). In gilthead sea bream, a high level of synchronicity between physical activity index (Fig. 5A) and respiratory frequency (Fig. 5B) was found, as evidenced by the positive correlation of extracted data (Fig. 5C). By contrast, European sea bass exhibited quite different activity patterns for physical activity index (Fig. 5D) and respiratory frequency (Fig. 5E), which showed a maximum value on the afternoon (acrophase at 18:04 h). This yielded an overall negative correlation between physical activity and respiration during almost all the recording time (Fig. 5F). Dynamics of recorded parameters also highlighted species-specific differences on the amplitude of the adjusted curves, which were 1.5- (physical activity) or 2-fold (respiratory frequency) higher in European sea bass than in gilthead sea bream.

*Table 1: Parameters of operculum velocity averaged lognormal for exercised gilthead sea bream and European sea bass. Values are the mean ± SD of the device for 2 min raw data measures from operculum movements of 18 gilthead sea bream individuals and 15 European sea bass individuals. For a given fish species and parameter, different superscript letters reflect significant (P < 0.001) differences with swimming speed. For a given swimming speed and parameter, asterisk reflects significant (P < 0.001) differences between fish species. For a given fish species and swimming speed, italics in $t_2 - t_1$ reflect significant (P < 0.001) differences with the corresponding $t_3 - t_2$.*

| | Gilthead sea bream | | | European sea bass | | |
|---|---|---|---|---|---|---|
| | 1 BL/s | 2 BL/s | 3 BL/s | 1 BL/s | 2 BL/s | 3 BL/s |
| $t_3 - t_1$ | $0.345 \pm 0.102^{a,*}$ | $0.257 \pm 0.095^{b,*}$ | $0.183 \pm 0.067^{c,*}$ | $0.390 \pm 0.096^{a}$ | $0.302 \pm 0.093^{b}$ | $0.208 \pm 0.067^{c}$ |
| $t_2 - t_1$ | $0.146 \pm 0.036^{a,*}$ | $0.112 \pm 0.035^{b,*}$ | $0.083 \pm 0.026^{c,*}$ | $0.163 \pm 0.034^{a}$ | $0.130 \pm 0.035^{b}$ | $0.093 \pm 0.027^{c}$ |
| $t_3 - t_2$ | $0.199 \pm 0.067^{a,*}$ | $0.145 \pm 0.060^{b,*}$ | $0.100 \pm 0.041^{c,*}$ | $0.227 \pm 0.062^{a}$ | $0.172 \pm 0.058^{b}$ | $0.115 \pm 0.041^{c}$ |
| Skew | $0.280 \pm 0.065^{a,*}$ | $0.221 \pm 0.065^{b,*}$ | $0.167 \pm 0.049^{c,*}$ | $0.309 \pm 0.062^{a}$ | $0.251 \pm 0.063^{b}$ | $0.186 \pm 0.049^{c}$ |
| Kurt | $3.147 \pm 0.079^{a,*}$ | $3.094 \pm 0.067^{b,*}$ | $3.054 \pm 0.041^{c,*}$ | $3.177 \pm 0.073^{a}$ | $3.120 \pm 0.062^{b}$ | $3.066 \pm 0.039^{c}$ |
| $\mu$ | $-0.476 \pm 0.050^{a,*}$ | $-0.526 \pm 0.048^{b,*}$ | $-0.571 \pm 0.035^{c,*}$ | $-0.453 \pm 0.049^{a}$ | $-0.500 \pm 0.052^{b}$ | $-0.555 \pm 0.039^{c}$ |
| $\sigma$ | $0.093 \pm 0.021^{a,*}$ | $0.073 \pm 0.021^{b,*}$ | $0.056 \pm 0.016^{c,*}$ | $0.102 \pm 0.020^{a}$ | $0.083 \pm 0.021^{b}$ | $0.062 \pm 0.016^{c}$ |
| D | $0.245 \pm 0.002^{a,*}$ | $0.290 \pm 0.002^{b,*}$ | $0.313 \pm 0.002^{c,*}$ | $0.351 \pm 0.002^{a}$ | $0.363 \pm 0.002^{b}$ | $0.434 \pm 0.004^{c}$ |

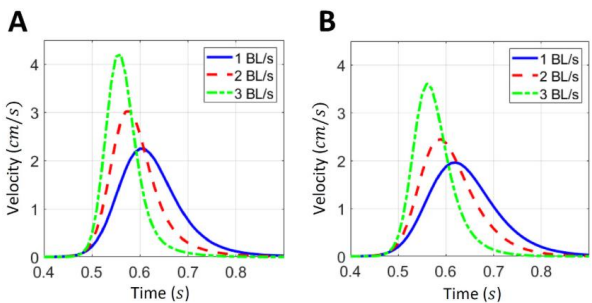

*Fig. 4. Average velocity lognormals of operculum movement. A. gilthead sea bream. B. European sea bass. Lognormals are represented at 1 BL/s (continuous blue), 2 BL/s (discontinuous red) and 3 BL/s (discontinuous green). (For interpretation of the references to color in this figure legend, the reader is referred to the web version of this article).*

Table 2 : Values of body movement velocity averaged lognormal for exercised gilthead sea bream and European sea bass. Values are the mean ± SD of the device for 2 min raw data measures from body tail movements of 18 gilthead sea bream individuals and 15 European sea bass individuals. For a given fish species and parameter, different superscript letters reflect significant (P < 0.001) differences with swimming speed. For a given swimming speed and parameter, asterisk reflects significant (P < 0.001) differences between fish species. For a given fish species and swimming speed, italics in $t_2 - t_1$ reflect significant (P < 0.001) differences with the corresponding $t_3 - t_2$.

| | Gilthead sea bream | | | European sea bass | | |
|---|---|---|---|---|---|---|
| | 1 BL/s | 2 BL/s | 3 BL/s | 1 BL/s | 2 BL/s | 3 BL/s |
| $t_3 - t_1$ | $0.387 \pm 0.135^{a,*}$ | $0.290 \pm 0.105^{b,*}$ | $0.209 \pm 0.083^{c,*}$ | $0.416 \pm 0.126^{a}$ | $0.309 \pm 0.090^{b}$ | $0.231 \pm 0.073^{c}$ |
| $t_2 - t_1$ | $0.160 \pm 0.046^{a,*}$ | $0.125 \pm 0.039^{b,*}$ | $0.093 \pm 0.033^{c,*}$ | $0.171 \pm 0.043^{a}$ | $0.132 \pm 0.036^{b}$ | $0.102 \pm 0.029^{c}$ |
| $t_3 - t_2$ | $0.227 \pm 0.089^{a,*}$ | $0.166 \pm 0.066^{b,*}$ | $0.116 \pm 0.050^{c,*}$ | $0.245 \pm 0.083^{a}$ | $0.177 \pm 0.062^{b}$ | $0.129 \pm 0.045^{c}$ |
| Skew | $0.310 \pm 0.090^{a,*}$ | $0.247 \pm 0.076^{b,*}$ | $0.188 \pm 0.063^{c,*}$ | $0.329 \pm 0.083^{a}$ | $0.261 \pm 0.070^{b}$ | $0.205 \pm 0.056^{c}$ |
| Kurt | $3.186 \pm 0.113^{a,*}$ | $3.119 \pm 0.078^{b,*}$ | $3.070 \pm 0.051^{c,*}$ | $3.206 \pm 0.105^{a}$ | $3.130 \pm 0.072^{b}$ | $3.080 \pm 0.046^{c}$ |
| $\mu$ | $-0.473 \pm 0.059^{a,*}$ | $-0.520 \pm 0.049^{b,*}$ | $-0.561 \pm 0.039^{c,*}$ | $-0.459 \pm 0.058^{a}$ | $-0.512 \pm 0.048^{b}$ | $-0.552 \pm 0.039^{c}$ |
| $\sigma$ | $0.103 \pm 0.029^{a,*}$ | $0.082 \pm 0.025^{b,*}$ | $0.062 \pm 0.021^{c,*}$ | $0.109 \pm 0.027^{a}$ | $0.086 \pm 0.023^{b}$ | $0.068 \pm 0.018^{c}$ |
| D | $0.263 \pm 0.003^{a,*}$ | $0.334 \pm 0.003^{b,*}$ | $0.362 \pm 0.003^{c,*}$ | $0.252 \pm 0.003^{a}$ | $0.214 \pm 0.003^{b}$ | $0.259 \pm 0.004^{a}$ |

Table 3: Pearson correlation coefficients between individual body weight and physical activity and respiratory frequency indexes in gilthead sea bream and European sea bass. P-value obtained in Pearson correlation is indicated in parentheses.

| | Gilthead sea bream | European sea bass |
|---|---|---|
| Physical activity index | −0.717 (0.109) | −0.447 (0.050) |
| Respiratory frequency | 0.811 (0.267) | 0.613 (0.106) |

## 4. Discussion

Traits related to locomotor performance and metabolism are subjected to natural selection as they are often coupled with important behaviors, such as predator evasion, prey capture, reproduction, migration and dominance (Boel et al., 2014; Killen et al., 2014a,b; Seebacher et al., 2013; Walker et al., 2005; Wilson et al., 2013). Moreover, an organism may specialize in one trial at the cost of the other, in which case the trade-off between antagonistic traits evolve causing phenotypic differentiation (Herrel et al., 2009; Heitz, 2014; Walker and Caddigan, 2015; Zhang et al., 2017). In fish, a good example is the trade-off between endurance capacity and sprint speed (Langerhans, 2009; Oufiero et al., 2011), and fish species with "active" lifestyle often have higher rates of dispersal in comparison to sedentary species (Réale et al., 2010; Careau and Garland, 2012). Selection for hypoxia resilience can also co-evolve with faster activity and increased dispersal ratios (Sinclair et al., 2014; Stoffels, 2015), which can be considered a positive trait in wildlife but not in farming conditions where animals cannot escape of deleterious oxygen conditions. Therefore, as reviewed by Davison and Herbert (2013) variations in swimming performance and behavior are of relevance to move towards a more precise and sustainable aquaculture production, although models of fish bioenergetics in swimming metabolic chambers are sometimes not easy to extrapolate to natural or rearing conditions. Indeed, metabolic rates of fish are different when they are moving in linear or nonlinear mode (Steinhausen et al., 2010), and interestingly we have observed that jerk accelerations of free-swimming fish at routine speed are apparently higher than those found for forced exercised fish in swim metabolic chambers (Martos-Sitcha et al., 2019). Besides, in-depth analysis of accelerometer records (raw data analysis) also provides valuable information about the amplitude and frequency of the operculum and body tail movements, helping to better phenotype the interspecies differences in locomotor capabilities (see below).

Most individual and species variations in locomotor performance and metabolism are coupled and linked to natural evolution as part of the complex behavioral ecology in a predator–prey system (Berger, 2010; Dias et al., 2018). Hence, European sea bass exhibits several morphological and physiological adaptations to sustain its lifestyle as a "fast" swimming predator (Spitz et al., 2013) with a spindle-shaped body that will reduce the water mass moving laterally at each tail stroke. Certainly, hydrodynamic models indicate that, when fish swimming is powered by fast white muscle fibers, muscle contractions are not only faster but possibly of larger amplitude than during slow muscle-powered cruising (Shadwick et al., 2013; Bale et al., 2015). Thus, comparing exercised European sea bass and gilthead sea bream, we found herein that the amplitude of body tail movements was larger and more stable in European sea bass, without apparent changes at lowmedium exercise intensity levels.

This might support higher speed accelerations and decelerations as characteristic features of a typical "fast" swimming predator. Indeed, in free-swimming fish, the range of variation of physical activity was also higher in European sea bass, though the average of jerk accelerations on a daily basis was lower in European sea bass than in gilthead sea bream

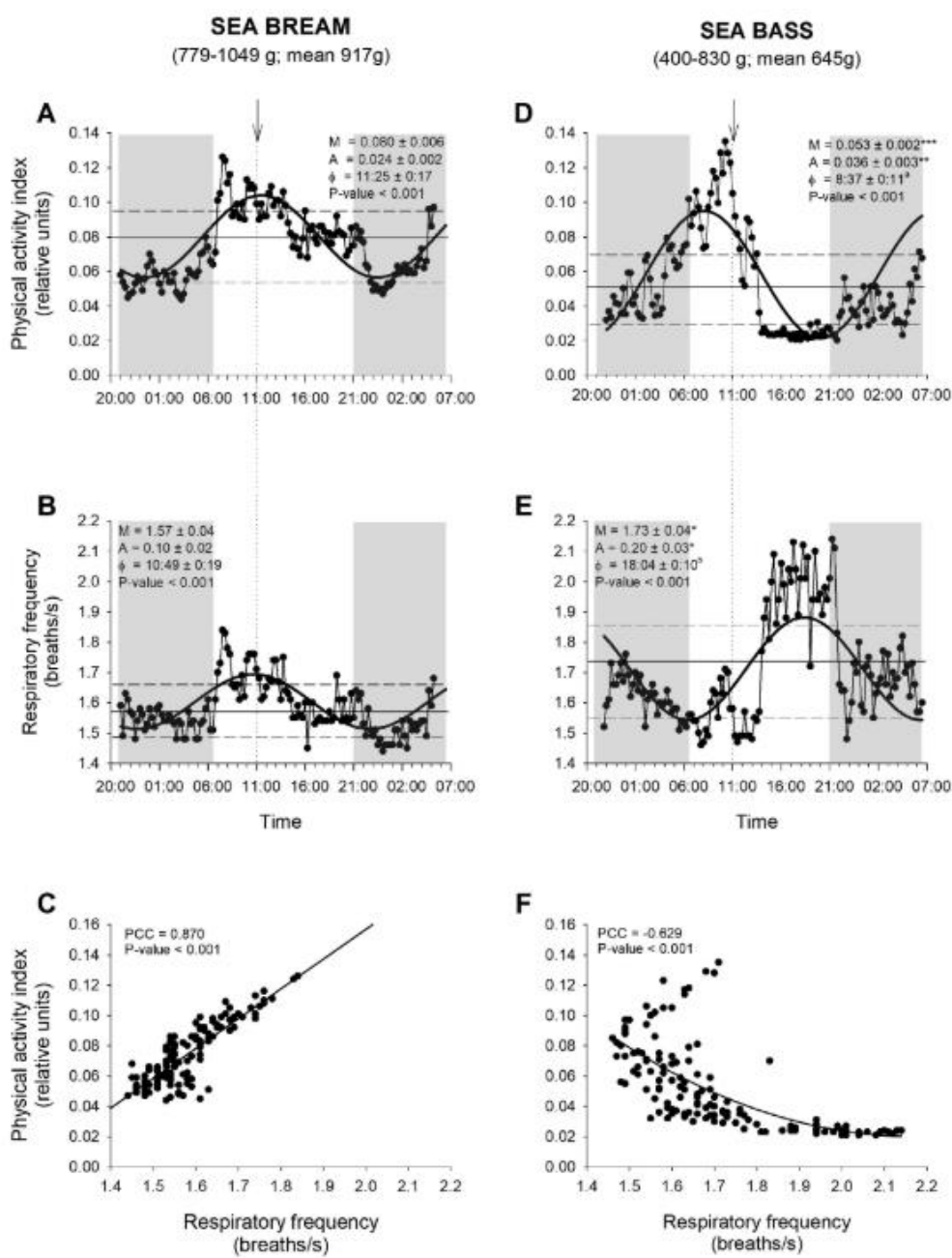

*Fig. 5. Consensus activity patterns from AEFishBIT measures. Daily variation of physical activity index (A, sea bream; D, sea bass) and respiratory frequency (B, sea bream; E, sea bass) in unfed free-swimming fish. At each time point, the mean value of 8 individuals is represented. Mesor is represented by a solid horizontal line, and dashed horizontal lines represent the 20 and 80 percentiles of mean time points. Gray shading represents the dark phase of the light cycle. Arrow marks feeding time (11:00 h) during the pre-recording period. The best-fit curves derived by cosinor analysis are shown as solid lines. Values of mesor (M), amplitude (A) and acrophase (Φ) are stated for each curve. Values represent mean ± SEM (n = 8). Asterisks indicate significant differences between species (\*P < 0.05, \*\*P < 0.01, \*\*\*P < 0.001; Student's t-test), and letters indicate significant differences between both AEFishBIT parameters in a same species (P < 0.05; Student's t-test). C, F. Correlation plots for a given time point between physical activity index and respiratory frequency in sea bream (C) and sea bass (F).*

Measurements of oxygen consumption are considered good indicators of the energy spent by fish to integrate a wide-range of biological processes, including the stress behavior under different challenging environments (Plaut, 2001; Remen et al., 2016). Thus, direct or indirect measurements of oxygen consumption (e.g. respiratory frequency) are of importance for underlining the metabolic scope of an individual. In this regard, it must be noticed that AEFishBIT calibration in Martos-Sitcha et al. (2019) elicited a close parallelism between measurements of oxygen consumption and respiratory frequency not only during moderate exercise, but also through the anaerobic phase that is largely increased at submaximal exercise (Ejbye-Ernst et al., 2016). As a general statement, we also reported in the first AEFishBIT study that European sea bass exhibits, in comparison to gilthead sea bream, a lower respiration at a given swimming speed, which was viewed as a better adaptation to fast swimming. This assumption was reinforced herein by the observation that a lower frequency of operculum movement was associated with a larger aperture, which in turn was more regulated than the frequency of movement with the increase of swimming speed. All these findings agree with the notion that operculum beats are a reliable measure of metabolic condition and locomotor capabilities in fish having buccal pumping as a mode of ventilation. However, its relevance is certainly limited in those species (e.g. tuna, sharks) that alternate buccal pumping with ram ventilation for covering their high oxygen demand during extreme exercise events (Brill and Bushnell, 2001; Wegner et al., 2010).

It is well known that exercise can profoundly influence the circadian system in rodents (Marchant et al., 1997; Mistlberger et al., 1997; Bobrzynska and Mrosovsky, 1998). In humans, there is also compelling evidence that exercise can elicit significant phase-shifting effects (Van Reeth et al., 1994; Edwards et al., 2002; Buxton et al., 2003) that facilitate the re-entrainment to a shifted light–dark and sleep–wake cycle (Miyazaki et al., 2001; Barger et al, 2004; Yamanaka et al., 2014). Activity patterns are also highly influenced by food availability, and the higher activity of white sea bream (Diplodus sargus) during the night in protected areas and artificial reefs is interpreted as the result of a tradeoff between predation risk and foraging needs (D'Anna et al., 2011). Early studies in gilthead sea bream also indicate that feeding time (scheduled vs. random) affects the behavior and physiology of the animal, and a single daily feeding cycle results beneficial because fish can prepare themselves for the forthcoming feed (Sánchez et al., 2009). Furthermore, in European sea bass and in a lower extent in gilthead sea bream, the percentage of individuals with a diurnal or nocturnal feeding behavior follows a dynamic cycle, which contributes to elicit the dual phasing behavior of some species to cope with anticipatory feed responses and seasonal changes in their environment (SánchezVázquez et al., 1998; Azzaydi et al., 2007; Vera et al., 2013). This also involves daily and seasonal cycles of hormonal activity, synchronized by the light-darkness and feeding-fasting cycles that enable different tissues to act as internal pacemakers (Isorna et al., 2017; Pérez-Sánchez et al., 2018). Certainly, the flexibility of the fish circadian and seasonal system makes these vertebrates a very interesting model for studying the communication between different functional oscillators. However, this relationship is often more complex than initially envisaged, and for instance zebrafish studies revealed an independent phasing between locomotor and feeding activities, which supports the concept of multioscillatory control of circadian rhythmicity in fish (Del Pozo et al., 2011). Thus, in our experimental setup, gilthead sea bream and European sea bass were fed once in the morning (11:00 h) during the prerecording period, but an anticipatory feed response by measures of physical activity was only especially evident in European sea bass, which might be favored by its lifestyle as a "fast" swimming predator.

In any case, the daily cycles of activity and respiration are out of phase in European sea bass, whereas they appeared highly synchronized in gilthead sea bream. This is indicative that swimming is largely fueled by aerobic metabolism in gilthead sea bream, but not in European sea bass that shares in the experimental conditions of the present study a more explosive swimming that would be mostly supported by the anaerobic white muscle fibers. From an energetic point of view, this type of behavior has an important impact on the net energy balance that contributes to explain the bad performance of European sea bass in comparison to gilthead sea bream across the production cycle (Torrecillas et al., 2017; Simó-Mirabet et al., 2018). However, regardless of these different metabolic features, correlation analysis in both European sea bass and gilthead sea bream support that selection for adult body size also selects for enhanced respiration and low activity. In other words, rearing conditions in our experimental facilities prime a phenotypic differentiation between fast growth and low activity and its antagonistic trait (slow growth and high activity) that are highly co-evolved through the evolution of modern teleosts (Rosenfeld et al., 2015; Sibly et al., 2015). From a practical point of view, this indicates that the enhanced energy cost of growth and maintenance is probably supported by a higher feed intake and perhaps improved feed conversion, as a result of a reduced locomotor activity that does not offer a special advantage in a scenario of intensive production under poorly restricted feeding (Devlin et al., 2004; Killen et al., 2014a,b). Nevertheless, it remains to be established the threshold level of physical activity to assure an active feeding behavior supporting fast growth.

In summary, the present study provides novel insights about the use of AEFishBIT for its use as a reliable tool for remote and individual sensing of fish behavior and metabolic status. It is designed to be attached to the operculum for recording at the same time fish accelerations and respiratory frequency (two in one) as an indicator of intraand inter-individual fish species differences in the energy portioning between growth and locomotor activities. The achieved results are supported by adaptive changes in body shape and specialized body movements, as a clear evidence that remote and individual monitoring of fish behavior can be used for recognizing beneficial behavioral patterns, which will allow researchers and farmers to select the most convenient lifestyles patterns, to establish stricter criteria of welfare and to improve rearing conditions for a more sustainable and ethical fish production.